\documentclass[runningheads]{llncs}

\usepackage{subcaption}
\usepackage{tabularx}

\usepackage{amssymb, amsmath, amsfonts}
\usepackage{acro}
\DeclareAcronym{ukbb}{
short=UKBB,
long=UK Biobank
}

\DeclareAcronym{mlp}{
  short=MLP,
  long=Multi-Layer Perceptron
  }

\DeclareAcronym{gcn}{
  short=GCN,
  long=Graph Convolutional Network
  }

\DeclareAcronym{gnn}{
  short=GNN,
  long=Graph Neural Network
  }

\DeclareAcronym{gat}{
  short=GAT,
  long=Graph Attention Network
  }
\usepackage{graphicx}
\usepackage{hyperref, xcolor}
\usepackage{booktabs}
\usepackage{array, multirow}

\usepackage[misc,geometry]{ifsym} 

\begin{document}

\title{A Comparative Study of Population-Graph Construction Methods and Graph Neural Networks for Brain Age Regression}

\titlerunning{Population-graphs and GNNs for Brain Age Regression}
\author{Kyriaki-Margarita Bintsi \inst{1} \textsuperscript{(\Letter)}, Tamara T. Mueller\inst{2},
Sophie Starck\inst{2}, \\ Vasileios Baltatzis \inst{1, 3}, Alexander Hammers \inst{3}, Daniel Rueckert \inst{1,2}}
\authorrunning{K. M. Bintsi et al.}

\institute{BioMedIA, Department of Computing, Imperial College London, UK \\
\email{m.bintsi19@imperial.ac.uk}\\ 
\and Lab for AI in Medicine and Healthcare, Faculty of Informatics, Technical University of Munich, Germany \\
\and Biomedical Engineering and Imaging Sciences, King's College London, UK 
}

\maketitle        

\begin{abstract}
The difference between the chronological and biological brain age of a subject can be an important biomarker for neurodegenerative diseases, thus brain age estimation can be crucial in clinical settings. One way to incorporate multimodal information into this estimation is through population graphs, which combine various types of imaging data and capture the associations among individuals within a population. In medical imaging, population graphs have demonstrated promising results, mostly for classification tasks. In most cases, the graph structure is pre-defined and remains static during training. However, extracting population graphs is a non-trivial task and can significantly impact the performance of \acp{gnn}, which are sensitive to the graph structure. In this work, we highlight the importance of a meaningful graph construction and experiment with different population-graph construction methods and their effect on \ac{gnn} performance on brain age estimation. We use the homophily metric and graph visualizations to gain valuable quantitative and qualitative insights on the extracted graph structures. For the experimental evaluation, we leverage the UK Biobank dataset, which offers many imaging and non-imaging phenotypes.
Our results indicate that architectures highly sensitive to the graph structure, such as \ac{gcn} and \ac{gat}, struggle with low homophily graphs, while other architectures, such as GraphSage and Chebyshev, are more robust across different homophily ratios. We conclude that static graph construction approaches are potentially insufficient for the task of brain age estimation and make recommendations for alternative research directions.
\end{abstract}
\acresetall
\section{Introduction}
Alzheimer's disease \cite{davatzikos2011prediction}, Parkinson's disease \cite{reeve2014ageing}, and schizophrenia \cite{koutsouleris2014accelerated}, among other neurodegenerative diseases, cause an atypically accelerated aging process in the brain. Consequently, the difference between an individual's biological brain age and their chronological age can act as an indicator of deviation from the normal aging trajectory \cite{alam2014morphological}, and potentially serve as a significant biomarker for neurodegenerative diseases \cite{cole2017predicting,franke2019ten}.

\acp{gnn} have been recently explored for medical tasks, since graphs can provide an inherent way of combining multi-modal information, and have demonstrated improved performance in comparison to graph-agnostic deep learning models \cite{parisot2018disease,ahmedt2021graph}. In most cases, the whole population is represented in a graph, namely population-graph, and the structure of the graph is pre-decided and remains static throughout the training. However, since the structure is not given but has to be inferred from the data, there are multiple ways that a graph can be constructed, which will possibly lead to different levels of performance. Every subject of the cohort is a node of the graph, with the imaging information usually allocated as node features. The interactions and relationships among the subjects of the cohort are represented by the edges. Two nodes are chosen to be connected based on a distance measure, or similarity score, usually based on the non-imaging phenotypes. The population graph is used as input to a \ac{gnn} for the task of node prediction, most commonly node classification. 

Brain MRI data have been widely used in population graphs in combination with \acp{gnn}. Parisot et al. were the first to propose a way to construct a static population graph using a similarity score that takes into account both imaging and non-imaging data \cite{parisot2018disease}. Many works were based on this work and extended the method to other applications \cite{kazi2019inceptiongcn,zhao2019graph}. When it comes to brain age regression, which is an inherently more complicated task than classification, there is little ongoing work using population graphs. To our knowledge, only Stankevičiūtė et al. \cite{stankeviciute2020population}  worked on the construction of the static population graph using the non-imaging information, but the predictive performance was relatively low.

In all of the cases mentioned above, the graph is chosen based on some criterion before training and remains static, thus it cannot be changed. However, since there are multiple ways that it can be constructed, there is no way to ensure that the final structure will be optimized for the task. This problem has been identified in the literature and multiple metrics have been described in order to evaluate the final graph structure \cite{zheng2022graph,ma2021homophily}, with homophily being the one most commonly used. A graph is considered homophilic, when nodes are connected to nodes with the same class label, and hence similar node features. Else, it is characterized as heterophilic. It has been found that some \acp{gnn}, such as \ac{gcn} \cite{kipf2016semi}, and \ac{gat} \cite{velivckovic2017graph}, are very sensitive to the graph structure, and a meaningless graph along with these models can perform worse than a graph-agnostic model. On the other hand, other models, such as GraphSage \cite{hamilton2017inductive} and Chebyshev \cite{defferrard2016convolutional}, are more resilient to the graph structure, and their performance is not affected as much by a heterophilic graph \cite{zhu2020beyond}.

In this work, we implement and evaluate the performance of different static graph construction methods for the task of brain age regression. We test their performance on the UK Biobank (UKBB), which offers a variety of both imaging and non-imaging phenotypes. The extracted population graphs are used along with the most popular \acp{gnn}, and more specifically \ac{gcn} \cite{kipf2016semi} and \ac{gat} \cite{velivckovic2017graph}, which are architectures highly sensitive to the graph structure, as well as GraphSage \cite{hamilton2017inductive} and Chebyshev \cite{defferrard2016convolutional}, which have been found to be resilient to the population-graph structure. The quantitative results and the visualization of the graphs allow us to draw conclusions and make suggestions regarding the use of static graphs for brain age regression. The code is available on GitHub at: \url{https://github.com/bintsi/brain-age-population-graphs} .

\section{Methods}
We have a dataset consisting of a set of $N$ subjects, each described by $M$ features. This dataset can be represented as $\textbf{X} = [\mathbf{x}_1, \ldots, \mathbf{x}_N] \in \mathbb{R}^{N \times M}$, where $\mathbf{x}_i$ represents the feature vector for the $i$-th subject. Additionally, each subject has a label denoted by $\textbf{y} \in \mathbb{R}^N$. Every subject $i$ is also characterized by a set of $K$ non-imaging phenotypes $\textbf{q}_i \in \mathbb{R}^K$. 

To establish the relationship between subjects, we introduce a population graph, denoted as $\mathcal{G}=\{\mathcal{V}, \mathcal{E}\}$. The graph consists of two components: $\mathcal{V}$ represents the set of nodes, where each subject corresponds to a unique node, and $\mathcal{E}$ represents the set of edges that define the connectivity between nodes.

In this paper, we explore four different graph construction approaches, which are described in detail below. In all of the graph construction methods, the node features consist of the imaging features described in the Dataset section \ref{subsec:dataset}.

\subsubsection{No edges}
As the baseline, we consider a graph with no edges among the nodes. This is the equivalent of traditional machine learning, where the node features act as the features used as input in the model. The machine learning model in this case is a \ac{mlp}. We refer to this approach as \textit{"No edges"}.

\subsubsection{Random Graph}
With the next approach, we want to explore whether the way of constructing the graph plays an important role on the performance of the \ac{gnn}. Thus, we build a random Erdos-Renyi graph \cite{erdHos1960evolution}, where a random number of nodes is chosen as neighbors for every node. 

\subsubsection{Clinical similarity score (Stankevičiūtė et al.)}
An approach of creating a population graph specifically for the task of brain age regression was proposed by \cite{stankeviciute2020population}, and it does not include the imaging features at all in the extraction of the edges. Instead, the edges are decided only from the non-imaging information.
More specifically, the similarity function for two subjects $i$ and $j$ is given by:
\begin{equation}
    sim(i, j) =  \frac{1}{K} \sum_{k=1}^{K}\boldsymbol{1} [q_{ik} = q_{jk}],
    \label{eq:kamilest}
\end{equation}
where  $q_{ik}$ is the value of the $k$-th non-imaging phenotype for the $k$-th subject, and $\boldsymbol{1}$  is the Kronecker delta function. Intuitively, the Kronecker delta function will only return 1 if the values of a particular non-imaging phenotype of two subjects match. Two nodes $i$ and $j$ are connected if $sim(i, j) \geq \mu$, with $\mu$ being a similarity threshold decided empirically. 


\subsubsection{Similarity score (Parisot et al.)}
In most of the related works in the literature, the way of creating the adjacency matrix $W$ was originally suggested by \cite{parisot2018disease} and it is calculated by:
\begin{equation}
    W(i,j) = Sim(\textbf{x}_i, \textbf{x}_j) \sum_{k=1}^{K} \gamma (q_{ik}, q_{jk}),
    \label{eq:parisot}
\end{equation}
where $Sim(\textbf{x}_i, \textbf{x}_j)$ is a similarity measure between the node features $(\textbf{x}_i, \textbf{x}_j)$ of the subjects $i$ and $j$, in our case cosine similarity, $\gamma ( \cdot, \cdot)$ is the distance of the non-imaging phenotypes between the nodes, and $q_{ik}$ is the value of the $k$-th non-imaging phenotype for the $i$-th subject.

The two terms in Eq. \ref{eq:parisot} indicate that both imaging, and non-imaging information are taken into account for the extraction of the edges. The second term is similar to the term $sim(i, j)$ of Eq. \ref{eq:kamilest}.

The computation of $\gamma ( \cdot, \cdot)$ is different for continuous and categorical features. For categorical data, $\gamma ( \cdot, \cdot)$ is defined as the Kronecker delta function $\textbf{1}$, as before.  For continuous data, $\gamma ( \cdot, \cdot)$ is defined as a unit-step function with respect to a specific threshold $\theta$. Intuitively, this means that the output of the $\gamma$ function will be 1 in case the continuous phenotypes of the two nodes are similar enough.


\subsubsection{kNN graph}
The last two graph construction approaches are based on the Nearest Neighbors (NN) algorithm. We connect the edges based on a distance function, in this case, cosine similarity. Each node is connected to its 5 closest neighbors. 

In the first approach, we use the neuroimaging information for the node features, and we also use the node features in order to estimate the distances of the nodes. We refer to this approach as \textit{"kNN (imaging)"}.

Similarly to before, in the second approach, we use the cosine similarity of a set of features in order to extract the graph structure. The node features incorporate the imaging information as before. The difference here is that we estimate the cosine similarity of the non-imaging phenotypes of the subjects with the purpose of finding the 5 closest neighbors. We refer to this approach as \textit{"kNN (non-imaging)"}.

Finally, we use all the available phenotypes, i.e. both the imaging, and non-imaging information, to connect the nodes, again using cosine similarity. We refer to this approach as \textit{"kNN (all phenotypes)"}.

\section{Experiments}
\subsection{Dataset}
\label{subsec:dataset}
For the experiments of the comparative study we use the \ac{ukbb} \cite{Alfaro-Almagro2018}, which not only offers an extensive range of vital organ images, including brain scans, but also contains a diverse collection of non-imaging information such as demographics, biomedical data, lifestyle factors, and cognitive performance measurements. Consequently, it is exceptionally well-suited for brain age estimation tasks that require integrating both imaging and non-imaging information.

To identify the most important factors influencing brain age in the \ac{ukbb}, we leverage the work of \cite{cole2020multimodality} and select 68 neuroimaging phenotypes and 20 non-imaging phenotypes that were found to be the most relevant to brain age in \ac{ukbb}. The neuroimaging features are obtained directly from the UKBB and include measurements derived from both structural MRI and diffusion-weighted MRI. 
All phenotypes are standardized to a normalized range between 0 and 1. 

The study focuses on individuals aged 47 to 81 years. We include only those subjects who have the necessary phenotypes available, resulting in a group of approximately 6500 subjects. The dataset is split into three parts: 75\% for training, 5\% for validation, and 20\% for testing.

\begin{table*}[t]
\caption{Performance of the different graph construction methods along with various \acp{gnn} on the test set. For every population graph, the homophily ratio is estimated. The baseline \ac{mlp} (\textit{No edges}) has a performance of MAE=3.73 years and and $R^2$ score of 0.56. Best performance for each \ac{gnn} model is highlighted in bold.}
\centering
\addtolength{\tabcolsep}{2pt}
\footnotesize
\begin{tabular}{lccccc}
\toprule
     \textbf{\multirow{3}{*}{\shortstack[l]{Graph \\ Construction}}} & \textbf{\ac{gcn}} &  \textbf{GraphSAGE} &  \textbf{\ac{gat}} & \textbf{Chebyshev} &\textbf{Homophily}\\
     \cmidrule{2-5}
     & \multicolumn{4}{c}{\textbf{MAE} (years)} &\\
     \cmidrule{2-5}
     Random graph & 5.19 &  3.72  & 5.38 & 3.83 & 0.7495 \\
     Parisot \cite{parisot2018disease} & 4.21  &  3.74 & 4.35 & 3.77 & 0.7899 \\
     Stankevičiūtė \cite{stankeviciute2020population} & 4.61 & 3.73  & 4.90 & 3.77  & 0.7743\\
     kNN (imaging) & \textbf{3.89} & 3.77 & 4.09 & 3.75 & 0.8259 \\
     kNN (non-imaging) &   4.76 & \textbf{3.68} & 4.98 & \textbf{3.72} & 0.7796 \\
     kNN (all phenotypes) & 3.93 & 3.76 & \textbf{4.07} & 3.73 & 0.8191\\
     & \multicolumn{4}{c}{\textbf{$R^2$ score}}&\\
     \cmidrule{2-5}
     Random graph & 0.26  & 0.59 & 0.2 & 0.54 \\
     Parisot \cite{parisot2018disease} & 0.49 &  0.59 & 0.47 & 0.55 \\
     Stankevičiūtė \cite{stankeviciute2020population} & 0.4  & 0.59  & 0.3 & 0.58\\
     kNN (imaging) & \textbf{0.56} & 0.58 & 0.51 & 0.59 \\
     kNN (non-imaging) & 0.38   & \textbf{0.59} &  0.3 & 0.55\\
     kNN (all phenotypes) &  0.56 & 0.58 & \textbf{0.53} & \textbf{0.6} \\
\bottomrule
\end{tabular} 
\label{tab:static_graphs_reg}
\end{table*}

\subsection{Results}
The graph construction methods described in the previous section are leveraged by four \ac{gnn} architectures and evaluated on the task of brain age regression. For every experiment, both spectral and spatial methods, and more specifically, \ac{gcn} \cite{kipf2016semi}, \ac{gat} \cite{velivckovic2017graph}, GraphSage \cite{hamilton2017inductive},  and Chebyshev \cite{defferrard2016convolutional}, are used. The results for all combinations of graph construction methods and \ac{gnn} architectures are shown in Table \ref{tab:static_graphs_reg}.

As a baseline, we use an \ac{mlp} (\textit{No edges}) that achieves a MAE of 3.73 years and a $R^2$ score of 0.56. For the case of \ac{gcn} and \ac{gat}, we notice that even though all of the graph costruction methods manage to extract a graph that outperforms the random graph, none of them manages to outperform the simple \ac{mlp}. The methods that used only the imaging information, i.e. the \textit{kNN graph (imaging)}, or all of the phenotypes, such as \textit{Parisot et. al \cite{parisot2018disease}}, extract graphs that when used as input to the \ac{gcn} or the \ac{gat} models, they perform better (MAE of around 4 years) compared to the ones that only use the non-imaging features for the extraction of the graph, such as \textit{kNN graph (non-imaging)} (MAE of 4.5-5 years). The GraphSAGE and Chebyshev models perform on similar levels regardless of the graph construction methods, with a MAE similar to the MAE given by the \ac{mlp}. The graph construction method that performs the best though, is the \textit{kNN graph (non-imaging)} along with the GraphSAGE, with a MAE of 3.68 years and a $R^2$ score of 0.59.

Finally, we estimate the homophily ratio of the population graphs extracted from the different construction methods based on \cite{mueller2023metrics} (Table \ref{tab:static_graphs_reg}). The most homophilic graphs tend to be the ones that use only imaging or a combination of imaging and non-imaging information for the extraction of the edges.


\subsection{Visualizations}
\begin{figure}[t]
\includegraphics[width=\textwidth]{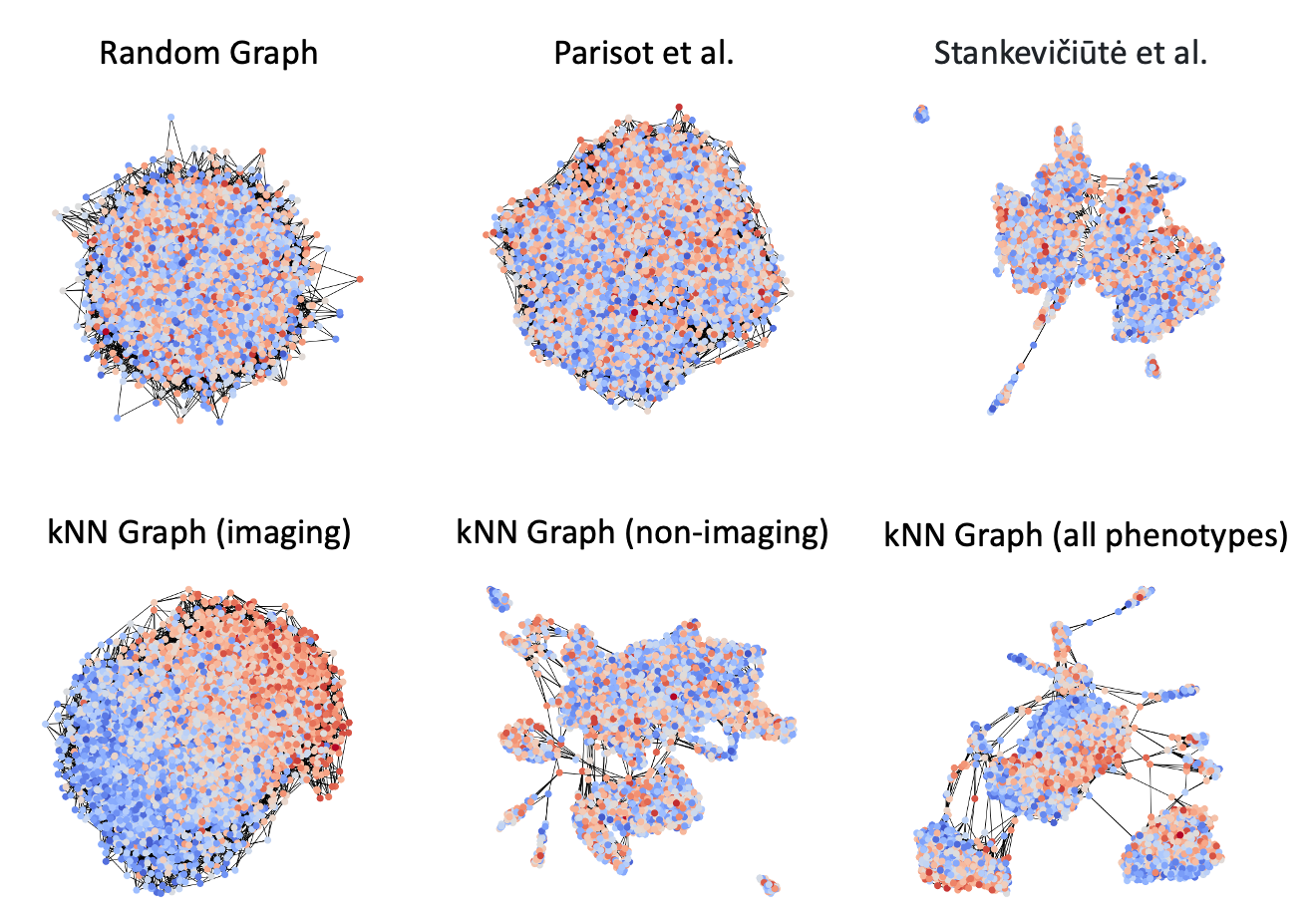}
\caption{Visualizations of the population graphs extracted from the different graph construction methods. The nodes are colored based on the age of the subject. Colder colors (i.e. blue) indicate subjects with an older age, while warmer colors (i.e. red) correspond to younger subjects.} \label{fig:graphs}
\end{figure}

Visualizing the extracted static graphs, can provide more insights about the performance of the different \ac{gnn} architectures. Therefore, we visualize the different population graphs colored based on the age of the subjects in Figure \ref{fig:graphs}. The various graph construction methods result in very different graph structures. More specifically, the graph construction methods that take into account the imaging phenotypes, either alone or in a combination with the non-imaging ones, provide graphs that are more meaningful, as subjects of similar ages are closer to each other. On the contrary, methods such as Stankevičiūtė et al. \cite{stankeviciute2020population}, create graphs that are similar to the random graph, with the neighborhoods not being very informative regarding the age of the subjects. 

\subsection{Implementation Details}
The static graphs extracted are chosen to be sparse for complexity reasons. The hyperparameters of the different construction methods were selected in such a way that all the extracted graphs have about 40000 to 50000 edges. Regarding the model hyperparameters, every \ac{gnn} contains a graph convolutional layer consisting of 512 units, followed by a fully connected layer with 128 units, prior to the regression layer. ReLU activation is chosen. The number of layers and their dimensions are determined by conducting a hyperparameter search based on validation performance. During training, the networks are optimized using the AdamW optimizer \cite{loshchilov2017decoupled} with a learning rate of 0.001 for 150 epochs, with the best model being saved. For the similarity threshold we use $\mu = 18$ and for the unit-step function threshold, we use $\theta = 0.1$. Both hyperparameters are selected based on validation performance and sparsity requirements. The implementation utilizes PyTorch Geometric \cite{Fey/Lenssen/2019} and a Titan RTX GPU. 

\section{Discussion and Conclusions}
In this work, we implement and evaluate static population graphs that are commonly used in the literature for other medical tasks, for brain age regression. We use the extracted graphs along with a number of popular \ac{gnn} models, namely \ac{gcn}, \ac{gat}, GraphSAGE, Chebyshev in order to get insights about the behavior of both the extraction methods, as well as the performance of the different models. By visualizing the graphs and estimating their homophily, we can provide further intuition in why the different static graphs do not work as expected and we highlight the problem that extracting static graphs from the data is not straightforward and possibly not suitable for brain age regression.

The reported results in Table \ref{tab:static_graphs_reg} indicate that the \ac{gcn} and \ac{gat} are highly sensitive to the graph structure, which is in agreement with the relevant literature \cite{zhu2020beyond,luan2022we}.
It is clear that the graph construction methods that provide graphs of higher homophily, lead to better performance for \ac{gcn} and \ac{gat} compared to the ones that provide a random-like population graph.

On the contrary, GraphSAGE and Chebyshev are not negatively affected by a graph structure with low homophily. This is because GraphSAGE encodes separately the node's encodings and the neighbors' encodings, which in our case are dissimilar, and hence it is affected less by the graph structure. When it comes to Chebyshev, the model is able to aggregate information from k hops in one layer, while the other \acp{gnn} achieve this through multiple layers. Being able to get information from higher order neighborhoods allows the model to find more relevant features, which would not be possible in the 1-hop neighborhood as this is highly heterophilic. But even these \acp{gnn}, that are more resilient to the graph structure, cannot distinctly outperform the \ac{mlp}. This behavior is expected as, according to the relevant literature, these models perform similarly to a \ac{mlp}, and they can outperform it only after a specific homophily threshold \cite{zhu2020beyond}. What is also observed is that the \textit{kNN (non-imaging)} works slightly better than the \ac{mlp} both for the GraphSAGE and the Chebyshev, probably because the models were able to capture the information added in the edges. 

To further explore the behavior of the graph construction methods and the population-graph they produce, we calculate the homophily ratio and we visualize the graphs. The graphs that are more homophilic perform better along with the models that are more sensitive to the graph structure, such as the \ac{gcn} and the \ac{gat}. We note here that all of the homophily ratios are higher than expected compared to the homophily reported in classification tasks, since we would expect that the random graph would have a homophily of 0.5. This is possibly because of the implementation of homophily for regression, as well as due to the imbalanced nature of the dataset. However, the trend is very clear and in agreement with the graph visualizations and the performance of the models.

All in all, the extraction of static population graphs for brain age regression, and in general for medical tasks for which the graph is not given, does not look very promising. In our opinion, there are multiple directions that should be explored. Firstly, one approach could be to learn the edges of the graph along with the training of the \ac{gnn}, which allows the extraction of an optimized graph for the specific task at-hand. There is some ongoing work on this adaptive graph learning \cite{wei2022graph,kazi2022differentiable,cosmo2020latent,bintsi2023multimodal}, but more focus should be given. In addition, the creation of \ac{gnn} models for graphs with high heterophily \cite{zheng2022graph,zhu2020beyond}, or the exploitation of graph rewiring techniques that could make the existing \acp{gnn} work better on heterophilic graphs \cite{bi2022make}, have proved to be useful for classification tasks. In the case of medical tasks, it might also be beneficial to incorporate the existing medical insights along with the above. It is also important to make the models and the graphs interpretable, as interpretability can be vital in healthcare, and it is something that is not currently widely explored in heterophilic graphs \cite{zheng2022graph}. Last but not least, introducing metrics that evaluate population graphs is of high importance, since this can help us understand better the structure of the graph. Even though there is some ongoing work when it comes to node classification \cite{luan2022we,zheng2022graph}, metrics regarding node regression have only recently started to being explored \cite{mueller2023metrics}.

\subsubsection{Acknowledgements} KMB would like to acknowledge funding from the EPSRC Centre for Doctoral Training in Medical Imaging (EP/L015226/1).


\bibliographystyle{splncs04}
\bibliography{bibliography}

\begin{thebibliography}{10}
\providecommand{\url}[1]{\texttt{#1}}
\providecommand{\urlprefix}{URL }
\providecommand{\doi}[1]{https://doi.org/#1}

\bibitem{ahmedt2021graph}
Ahmedt-Aristizabal, D., Armin, M.A., Denman, S., Fookes, C., Petersson, L.:
  Graph-based deep learning for medical diagnosis and analysis: past, present
  and future. Sensors  \textbf{21}(14), ~4758 (2021)

\bibitem{alam2014morphological}
Alam, S.B., Nakano, R., Kamiura, N., Kobashi, S.: Morphological changes of
  aging brain structure in mri analysis. In: 2014 Joint 7th International
  Conference on Soft Computing and Intelligent Systems (SCIS) and 15th
  International Symposium on Advanced Intelligent Systems (ISIS). pp. 683--687.
  IEEE (2014)

\bibitem{Alfaro-Almagro2018}
Alfaro-Almagro, F., Jenkinson, M., Bangerter, N.K., Andersson, J.L., Griffanti,
  L., Douaud, G., Sotiropoulos, S.N., Jbabdi, S., Hernandez-Fernandez, M.,
  Vallee, E., Vidaurre, D., Webster, M., McCarthy, P., Rorden, C., Daducci, A.,
  Alexander, D.C., Zhang, H., Dragonu, I., Matthews, P.M., Miller, K.L., Smith,
  S.M.: {Image processing and Quality Control for the first 10,000 brain
  imaging datasets from UK Biobank}. NeuroImage  \textbf{166},  400--424 (2
  2018)

\bibitem{bi2022make}
Bi, W., Du, L., Fu, Q., Wang, Y., Han, S., Zhang, D.: Make heterophily graphs
  better fit gnn: A graph rewiring approach. arXiv preprint arXiv:2209.08264
  (2022)

\bibitem{bintsi2023multimodal}
Bintsi, K.M., Baltatzis, V., Potamias, R.A., Hammers, A., Rueckert, D.:
  Multimodal brain age estimation using interpretable adaptive population-graph
  learning. arXiv preprint arXiv:2307.04639  (2023)

\bibitem{cole2020multimodality}
Cole, J.H.: Multimodality neuroimaging brain-age in uk biobank: relationship to
  biomedical, lifestyle, and cognitive factors. Neurobiology of aging
  \textbf{92},  34--42 (2020)

\bibitem{cole2017predicting}
Cole, J.H., Poudel, R.P., Tsagkrasoulis, D., Caan, M.W., Steves, C., Spector,
  T.D., Montana, G.: Predicting brain age with deep learning from raw imaging
  data results in a reliable and heritable biomarker. NeuroImage  \textbf{163},
   115--124 (2017)

\bibitem{cosmo2020latent}
Cosmo, L., Kazi, A., Ahmadi, S.A., Navab, N., Bronstein, M.: Latent-graph
  learning for disease prediction. In: Medical Image Computing and Computer
  Assisted Intervention--MICCAI 2020: 23rd International Conference, Lima,
  Peru, October 4--8, 2020, Proceedings, Part II 23. pp. 643--653. Springer
  (2020)

\bibitem{davatzikos2011prediction}
Davatzikos, C., Bhatt, P., Shaw, L.M., Batmanghelich, K.N., Trojanowski, J.Q.:
  Prediction of mci to ad conversion, via mri, csf biomarkers, and pattern
  classification. Neurobiology of aging  \textbf{32}(12),  2322--e19 (2011)

\bibitem{defferrard2016convolutional}
Defferrard, M., Bresson, X., Vandergheynst, P.: Convolutional neural networks
  on graphs with fast localized spectral filtering. Advances in neural
  information processing systems  \textbf{29} (2016)

\bibitem{erdHos1960evolution}
Erd{\H{o}}s, P., R{\'e}nyi, A., et~al.: On the evolution of random graphs.
  Publ. math. inst. hung. acad. sci  \textbf{5}(1),  17--60 (1960)

\bibitem{Fey/Lenssen/2019}
Fey, M., Lenssen, J.E.: Fast graph representation learning with {PyTorch
  Geometric}. In: ICLR Workshop on Representation Learning on Graphs and
  Manifolds (2019)

\bibitem{franke2019ten}
Franke, K., Gaser, C.: Ten years of brainage as a neuroimaging biomarker of
  brain aging: what insights have we gained? Frontiers in neurology p.~789
  (2019)

\bibitem{hamilton2017inductive}
Hamilton, W., Ying, Z., Leskovec, J.: Inductive representation learning on
  large graphs. Advances in neural information processing systems  \textbf{30}
  (2017)

\bibitem{kazi2022differentiable}
Kazi, A., Cosmo, L., Ahmadi, S.A., Navab, N., Bronstein, M.: Differentiable
  graph module (dgm) for graph convolutional networks. IEEE Transactions on
  Pattern Analysis and Machine Intelligence  (2022)

\bibitem{kazi2019inceptiongcn}
Kazi, A., Shekarforoush, S., Arvind~Krishna, S., Burwinkel, H., Vivar, G.,
  Kort{\"u}m, K., Ahmadi, S.A., Albarqouni, S., Navab, N.: Inceptiongcn:
  receptive field aware graph convolutional network for disease prediction. In:
  Information Processing in Medical Imaging: 26th International Conference,
  IPMI 2019, Hong Kong, China, June 2--7, 2019, Proceedings 26. pp. 73--85.
  Springer (2019)

\bibitem{kipf2016semi}
Kipf, T.N., Welling, M.: Semi-supervised classification with graph
  convolutional networks. arXiv preprint arXiv:1609.02907  (2016)

\bibitem{koutsouleris2014accelerated}
Koutsouleris, N., Davatzikos, C., Borgwardt, S., Gaser, C., Bottlender, R.,
  Frodl, T., Falkai, P., Riecher-R{\"o}ssler, A., M{\"o}ller, H.J., Reiser, M.,
  et~al.: Accelerated brain aging in schizophrenia and beyond: a
  neuroanatomical marker of psychiatric disorders. Schizophrenia bulletin
  \textbf{40}(5),  1140--1153 (2014)

\bibitem{loshchilov2017decoupled}
Loshchilov, I., Hutter, F.: Decoupled weight decay regularization. arXiv
  preprint arXiv:1711.05101  (2017)

\bibitem{luan2022we}
Luan, S., Hua, C., Lu, Q., Zhu, J., Chang, X.W., Precup, D.: When do we need
  gnn for node classification? arXiv preprint arXiv:2210.16979  (2022)

\bibitem{ma2021homophily}
Ma, Y., Liu, X., Shah, N., Tang, J.: Is homophily a necessity for graph neural
  networks? arXiv preprint arXiv:2106.06134  (2021)

\bibitem{mueller2023metrics}
Mueller, T., Starck, S., Feiner, L.F., Bintsi, K.M., Rueckert, D., Kaissis, G.:
  Extended graph assessment metrics for regression and weighted graphs. arXiv
  preprint  (2023)

\bibitem{parisot2018disease}
Parisot, S., Ktena, S.I., Ferrante, E., Lee, M., Guerrero, R., Glocker, B.,
  Rueckert, D.: Disease prediction using graph convolutional networks:
  application to autism spectrum disorder and alzheimer’s disease. Medical
  image analysis  \textbf{48},  117--130 (2018)

\bibitem{reeve2014ageing}
Reeve, A., Simcox, E., Turnbull, D.: Ageing and parkinson's disease: why is
  advancing age the biggest risk factor? Ageing research reviews  \textbf{14},
  19--30 (2014)

\bibitem{stankeviciute2020population}
Stankeviciute, K., Azevedo, T., Campbell, A., Bethlehem, R., Lio, P.:
  Population graph gnns for brain age prediction. In: ICML Workshop on Graph
  Representation Learning and Beyond (GRL+). pp. 17--83 (2020)

\bibitem{velivckovic2017graph}
Veli{\v{c}}kovi{\'c}, P., Cucurull, G., Casanova, A., Romero, A., Lio, P.,
  Bengio, Y.: Graph attention networks. arXiv preprint arXiv:1710.10903  (2017)

\bibitem{wei2022graph}
Wei, S., Zhao, Y.: Graph learning: A comprehensive survey and future
  directions. arXiv preprint arXiv:2212.08966  (2022)

\bibitem{zhao2019graph}
Zhao, X., Zhou, F., Ou-Yang, L., Wang, T., Lei, B.: Graph convolutional network
  analysis for mild cognitive impairment prediction. In: 2019 IEEE 16th
  International Symposium on Biomedical Imaging (ISBI 2019). pp. 1598--1601.
  IEEE (2019)

\bibitem{zheng2022graph}
Zheng, X., Liu, Y., Pan, S., Zhang, M., Jin, D., Yu, P.S.: Graph neural
  networks for graphs with heterophily: A survey. arXiv preprint
  arXiv:2202.07082  (2022)

\bibitem{zhu2020beyond}
Zhu, J., Yan, Y., Zhao, L., Heimann, M., Akoglu, L., Koutra, D.: Beyond
  homophily in graph neural networks: Current limitations and effective
  designs. Advances in Neural Information Processing Systems  \textbf{33},
  7793--7804 (2020)

\end{thebibliography}

\end{document}